\documentclass{article}

\PassOptionsToPackage{hyperfootnotes=false}{hyperref}
\usepackage[preprint]{paper/roboralscan_preprint}
\usepackage{graphicx}
\graphicspath{{figures/}{paper/figures/}}
\usepackage{amsmath}
\usepackage{amssymb}
\usepackage{subcaption}
\usepackage{booktabs} 
\usepackage{multirow} 
\usepackage{pifont}
\usepackage{placeins}
\hypersetup{
  pdftitle={RobOralScan: Learning Active Intraoral Scanning for Robotic Dental Reconstruction},
  pdfauthor={Jinhyung Lee, Haeun Yun, Siwon Kim, Gihyun Baek, Sungho Moon, Sehyun Hwang, Sunghoon Im}
}

\newcommand{\firstcontrib}{\textsuperscript{*}}
\newcommand{\secondcontrib}{\textsuperscript{\ensuremath{\dagger}}}
\newcommand{\corresponding}{\textsuperscript{\ensuremath{\ddagger}}}
\newcommand{\roboralscanauthornotes}{%
  \begingroup
  \renewcommand{\thefootnote}{*}
  \footnotetext{Equal contribution. The first four authors contributed equally to this work.}
  \endgroup
  \begingroup
  \renewcommand{\thefootnote}{\ensuremath{\dagger}}
  \footnotetext{Equal second contribution. The two authors contributed equally to this work.}
  \endgroup
  \begingroup
  \renewcommand{\thefootnote}{\ensuremath{\ddagger}}
  \footnotetext{Corresponding author: Sunghoon Im (\texttt{sunghoonim@dgist.ac.kr}).}
  \endgroup
}
\makeatletter
\newcommand{\maybeauthornotes}{%
  \if@conferencefinal
    \roboralscanauthornotes
  \else
    \if@preprinttype
      \roboralscanauthornotes
    \fi
  \fi
}
\makeatother

\title{RobOralScan: Learning Active Intraoral Scanning for Robotic Dental Reconstruction}

\author{
  Jinhyung Lee\firstcontrib\\
  DGIST\\
  \texttt{refracta@dgist.ac.kr}
  \And
  Haeun Yun\firstcontrib\\
  DGIST\\
  \texttt{haeun@dgist.ac.kr}
  \And
  Siwon Kim\firstcontrib\\
  DGIST\\
  \texttt{c12030@dgist.ac.kr}
  \And
  Gihyun Baek\firstcontrib\\
  DGIST\\
  \texttt{gh.baek@dgist.ac.kr}
  \AND
  Sungho Moon\secondcontrib\\
  DGIST\\
  \texttt{byeol3325@dgist.ac.kr}
  \And
  Sehyun Hwang\secondcontrib\\
  DGIST\\
  \texttt{sehyun030@gmail.com}
  \And
  Sunghoon Im\corresponding\\
  DGIST\\
  \texttt{sunghoonim@dgist.ac.kr}
}

\begin{document}
\maketitle
\maybeauthornotes

\begin{abstract}
Intraoral scanning is widely used for digital optical impressions in prosthodontic, implant, and orthodontic treatment, but full-arch and long-span scanning remain labor-intensive tasks with limited automation. In the confined oral cavity, operators must continuously adjust scanner motion while accumulating narrow field-of-view observations, making reconstruction quality sensitive to missing tooth surfaces and operator workload.
We propose RobOralScan, which, to the best of our knowledge, is the first reinforcement learning (RL)-based pipeline for robotic automatic intraoral scanning. RobOralScan introduces a geometric memory-based observation space that accumulates partial scan observations into a tri-state geometric representation, allowing the policy to reason over scan history and insufficiently observed regions. It further introduces tooth-wise coverage learning, combining coverage-aware reward signals and a progressive training scheme to improve global reconstruction coverage while reducing uneven coverage across individual teeth. The learned policy selects relative scanner motions from accumulated geometric memory and robot proprioception for closed-loop scan control within the oral workspace.
RobOralScan achieves a Chamfer Distance of $0.00838$, an average coverage of $92.58\%$, a lower-tail per-tooth coverage of $88.45\%$, and a normalized AUC of $0.6674$, completing the scan criterion in $8$ of $10$ evaluation episodes. Furthermore, zero-shot sim-to-real experiments demonstrate its practical feasibility on a physical robot-scanner setup.
\end{abstract}



\section{Introduction}
\label{sec:introduction}

Intraoral scanning has become an essential entry point for digital prosthodontic, implant, and orthodontic workflows. By replacing conventional physical impressions with digital optical impressions, it reduces patient discomfort, gag reflex, and chairside time~\citep{Yuzbasioglu2014DigitalConventional,Gjelvold2016DigitalConventional,Pachiou2025PROMs,RamosMorro2026PatientPerception}. It also provides three-dimensional intraoral data that can be directly integrated into computer-aided design, manufacturing, and treatment-planning pipelines. As digital dentistry increasingly depends on accurate three-dimensional intraoral data, the clinical need for reliable, repeatable, and efficient intraoral scanning continues to grow. 

However, complete-arch and long-span intraoral scans remain high-burden acquisition tasks for which automation has not yet become established. The operator should continuously observe teeth and adjacent gingiva through a narrow field of view inside the confined oral cavity while managing viewing distance and angle, motion speed, path repeatability, and missed regions. Such manual path control can affect final surface deviation and precision~\citep{Mangano2017IOSReview,Kwon2021FiveIOS,Fratila2025Guideline,Choi2024ScanPath,Schlogl2025ScanningPatterns}, and scan quality depends strongly on operator expertise and workload~\citep{Limones2025OperatorExperienceReview}. These limitations indicate that intraoral scanning is not only a clinically important digital acquisition step, but also a compelling target for automation to reduce operator-dependent variability and improve the consistency of downstream digital dental workflows.

More broadly, medical image acquisition is increasingly being reframed as an active, robot-assisted, and learning-based control problem, rather than a purely manual sensing procedure. In robotic ultrasound and endoscopic scanning, recent studies have investigated automated scan-path planning, probe or camera motion control, and patient-specific adaptation in anatomically constrained environments~\citep{bi2026autonomous,chen2025ultradp,zhang2022deep}. SonoGym~\citep{ao2025sonogym} further reflects this trend by formulating robotic ultrasound navigation, anatomy reconstruction, and surgical guidance as learnable simulation tasks, presenting a paradigm in which medical image acquisition itself can be treated as a robot policy learning problem. In contrast, robotic automatic intraoral scanning remains relatively underexplored due to the need for sequential scanner-pose adjustment within the confined oral workspace while jointly considering tooth-wise surface coverage and motion feasibility.

To address this gap, this paper proposes RobOralScan, which, to the best of our knowledge, is the first reinforcement learning (RL)-based robotic automatic dental scanning pipeline. Unlike fixed or manually designed scan-path procedures, RobOralScan formulates intraoral scanning as a feedback-driven sequential scan-control problem. At each step, the policy selects the next scanner motion from accumulated scan observations and robot state, allowing the scan path to adapt to insufficiently observed tooth surfaces while the robot arm provides repeatable scanner pose execution within the confined oral workspace.

The main technical contributions of RobOralScan are twofold. First, we propose a geometric memory-based observation space for RL-based intraoral scanning. Instead of relying only on the current scan frame, the policy observes an accumulated geometric representation that summarizes scan history and provides information about previously observed and insufficiently observed regions. This enables the policy to reason about coverage progress under partial observations. Second, we propose tooth-wise coverage learning, which combines coverage-aware reward signals and a progressive training scheme. This learning objective encourages the policy to improve overall reconstruction quality while reducing uneven coverage across individual teeth and avoiding inefficient or infeasible scanning behavior. Reconstructed point clouds are finally evaluated against reference surfaces using consistent geometric metrics. Representative results and ablations show that the proposed observation space and coverage-learning design improve reconstruction quality and coverage efficiency.
The contributions of this paper are as follows:

\begin{itemize}
    \item We introduce RobOralScan, which, to the best of our knowledge, is the first reinforcement learning (RL)-based pipeline for robotic automatic dental scanning.

    \item We propose a geometric memory-based observation space that accumulates partial scan observations into a tri-state representation, allowing the policy to reason about scan history and insufficiently observed tooth surfaces.

    \item We propose tooth-wise coverage learning, consisting of coverage-aware reward signals and a progressive training scheme, to improve global reconstruction coverage while reducing uneven coverage across individual teeth.

\end{itemize}


\section{Related Work}
\label{sec:related_work}

\textbf{Intraoral Scanning.}
Digital intraoral scanners are widely used for optical impressions in prosthodontic, implant, and orthodontic workflows, but complete-arch and long-span scans remain sensitive to scanner motion, scan strategy, and operator-dependent acquisition conditions~\cite{richert2017intraoral, alkadi2023comprehensive, hardan2023scanning}. Prior studies have shown that scan paths and scanning strategies affect the trueness, precision, and acquisition time of complete-arch intraoral scans~\cite{ender2013influence, medina2018accuracy, mai2022impact}. These studies indicate that scanner motion is a key factor in reconstruction quality, but they mainly analyze manually designed scan paths or operator-dependent protocols. 

\textbf{Next-Best-View and Active 3D Reconstruction.}
The Next-Best-View (NBV) problem studies how an agent should sequentially select sensor poses to acquire informative observations of an initially unknown scene~\cite{connolly1985determination, zeng2020view}. 
Early and classical approaches typically estimate information gain from geometric representations such as volumetric occupancy maps, allowing view candidates to be evaluated according to expected surface discovery or uncertainty reduction~\cite{isler2016information, delmerico2018comparison}. 
More recent learning-based methods formulate view planning as a sequential decision-making problem, often using Markov decision processes to learn policies that reduce reliance on manually designed heuristics~\cite{peralta2020next}. 
Subsequent work has further investigated generalization to unseen object categories and datasets~\cite{chen2024gennbv}, as well as objectives that more directly reflect reconstruction fidelity rather than coverage alone~\cite{frahm2025vin}. 
These studies establish NBV as a principled framework for active perception, but most formulations are designed around object-centric or relatively unconstrained sensing settings. 

\textbf{Robotic Medical Scanning.}
Robotic scanning has been actively studied in medical imaging, particularly in ultrasound, where autonomous probe control can reduce operator dependence and improve reproducibility. 
Learning-based approaches have demonstrated the potential of reinforcement learning policies trained in physics-based simulation for surgical and anatomical scanning tasks~\cite{ao2025sonogym, bi2026autonomous, ao2026robust}.
Recent policy-learning methods, including diffusion-based controllers, have also explored transfer to unseen subjects and more flexible scan execution~\cite{chen2025ultradp}. 
Beyond ultrasound, robot-guided multi-view reconstruction~\cite{gobel2025robot} and coverage-driven policies for capsule endoscopy~\cite{zhang2022deep} show how robotic motion planning can support visual acquisition in medical environments. 
These works are closely related in their use of closed-loop sensing and robot control, and they demonstrate the broader feasibility of autonomous medical scanning. 

Our work connects these three lines of research: active view planning provides the sequential perception framework, robotic medical scanning provides the closed-loop control perspective, and dental scan-strategy studies motivate tooth-wise acquisition objectives. 
Building on these insights, RobOralScan formulates intraoral scanning as a feedback-driven robotic scan-control problem tailored to the geometric and workspace constraints of dental acquisition.

\section{Method}
\label{sec:method}

\begin{figure}[t]
    \centering
    \includegraphics[width=\columnwidth]{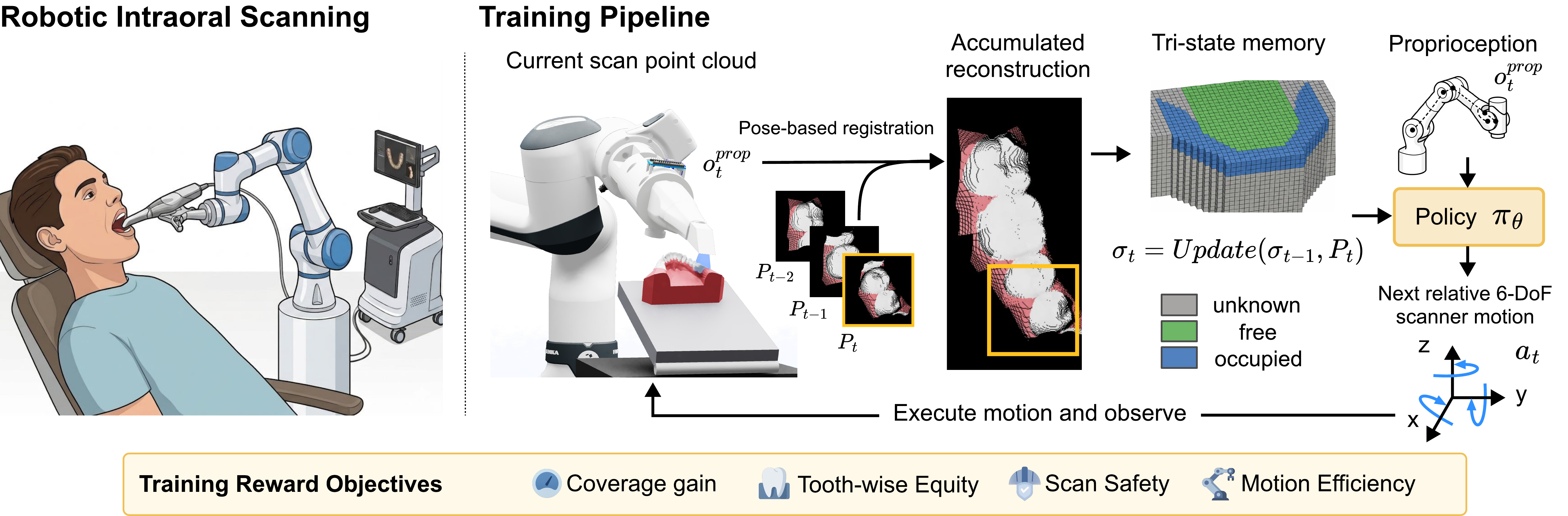}
    \caption{System overview of RobOralScan. Local scan observations are accumulated into a tri-state geometric memory, which is provided to the policy together with robot proprioception. The policy outputs relative 6-DoF scanner motions for closed-loop scan control.
}
    \label{fig:overview}
\end{figure}


RobOralScan introduces two main technical components: a geometric memory-based observation space and tooth-wise coverage learning. The geometric memory-based observation space enables the policy to reason over accumulated scan history described in Sec.~\ref{sec:formulation}. The tooth-wise coverage learning guides the policy toward both high global reconstruction coverage and balanced coverage across individual teeth described in Sec.~\ref{sec:coverage_learning}. An overview of the system is shown in Fig.~\ref{fig:overview}.



\subsection{Geometric Memory-based Observation Space for Sequential Scan Control}
\label{sec:formulation}
We formulate robotic intraoral scanning as a partially observed episodic reinforcement learning problem. Each episode corresponds to scanning a single patient jaw for at most $T$ steps. At each timestep, the scanner captures a point cloud $P_t \subset \mathbb{R}^3$ of the visible surface and updates the accumulated point cloud $\bar{P}_t = \bigcup_{\tau=0}^{t} P_\tau$. The agent observes $\mathbf{o}_t \in \mathcal{O}$ from the accumulated scan state and robot state, executes an action $\mathbf{a}_t \in \mathcal{A}$, and receives a scalar reward $r_t$. The underlying state $s_t \in \mathcal{S}$ is not directly available to the policy, as it includes the complete tooth geometry that is only partially revealed through scanning. The goal is to learn a policy that selects scanner motions to improve reconstruction coverage, reduce poorly covered teeth, and maintain feasible motion within the intraoral workspace. 

\textbf{Geometric memory-based observation.} The policy observation $\mathbf{o}_t = (\boldsymbol{\sigma}_t, \mathbf{o}^{\mathrm{prop}}_t)$ consists of an accumulated tri-state volumetric memory $\boldsymbol{\sigma}_t \in \mathcal{C}^{H \times W \times D}$, where $\mathcal{C} = \{\textit{unknown}, \textit{free}, \textit{occupied}\}$, and a robot proprioception vector $\mathbf{o}^{\mathrm{prop}}_t \in \mathbb{R}^{21}$.
Here, $H, W,$ and $D$ denote the dimensions of a fixed voxel grid $\mathcal{V}$ defined over the scanning workspace.

We construct the volumetric memory $\boldsymbol{\sigma}_t$ by assigning each individual voxel $v \in \mathcal{V}$ to one of the three states in $\mathcal{C}$. Let $\mathcal{V}^{\mathrm{occ}}_t = \{v \in \mathcal{V}\mid \exists\, p \in \bar{P}_t,\; p \in v\}$ denote the set of voxels containing at least one point after voxelizing $\bar{P}_t$ onto $\mathcal{V}$. The state of each voxel $\sigma_t(v)$, is then

\begin{equation}
\sigma_t(v) = \begin{cases}
\textit{occupied}, & \text{if } v \in \mathcal{V}^{\mathrm{occ}}_t, \\
\textit{free},     & \text{if } v \in \mathcal{R}_t \text{ and } v \notin \mathcal{V}^{\mathrm{occ}}_t, \\
\textit{unknown},  & \text{otherwise.}
\end{cases}
\end{equation}


where $\mathcal{R}_t \subset \mathcal{V}$ denote the set of voxels traversed by the scanner's line-of-sight rays up to time $t$. Following standard occupancy mapping techniques~\cite{hornung2013octomap}, we compute $\mathcal{R}_t$ via 3D ray-casting from the camera's optical center to the observed surface points. 
Intuitively, a voxel is \textit{occupied} if it contains at least one point from the accumulated point cloud $\bar{P}_t$ (i.e., a scanned tooth surface). It is \textit{free} if camera rays have passed through it without intersecting any surface. All other unobserved voxels default to \textit{unknown}. The boundary between \textit{occupied} and \textit{unknown} regions implicitly encodes the unexplored frontiers, providing the policy with coverage progress information.

\paragraph{Incremental scanner action.}
\label{sec:action}

The action $\mathbf{a}_t \in \mathbb{R}^{6}$ is defined as a relative scanner pose offset in the Tool Center Point frame:
\begin{equation}
\mathbf{a}_t =
(\delta x, \delta y, \delta z, \delta \phi, \delta \theta, \delta \psi)^\top .
\end{equation}
Here, $(\delta x, \delta y, \delta z)$ denotes the translational offset and $(\delta \phi, \delta \theta, \delta \psi)$ denotes the rotational offset. We use relative actions since intraoral scanning requires continuous local adjustments from the current scanner pose rather than generating a complete scan trajectory at once. The raw policy output is scaled, clamped to the valid workspace, and converted into robot joint targets through Differential Inverse Kinematics.

\subsection{Tooth-wise Coverage Learning}
\label{sec:coverage_learning}
We design the learning objective to encourage both overall reconstruction improvement and balanced per-tooth coverage. During training, a reference-surface coverage tracker with ground-truth tooth annotations computes coverage-based learning signals. Ground-truth annotations are used only for reward computation and are not included in the policy observation.

Let $c_{k,t}$ denote the coverage of tooth $k$ at timestep $t$. The tracker also returns total tooth surface coverage $c^{\text{teeth}}_t$ and tooth-adjacent gum coverage $c^{\text{adj-gum}}_t$. We define the global coverage score as
\begin{equation}
\bar{c}_t =
\tfrac{1}{2}
\left(
c^{\text{teeth}}_t +
c^{\text{adj-gum}}_t
\right).
\end{equation}
We use this score to compute global coverage progress in the reward and to report average coverage at evaluation time.

\paragraph{Coverage-aware reward signals.}
The total reward is composed of coverage completion, tooth-wise equity, scan safety, and motion efficiency terms:
\begin{equation}
r_t =
r^{\mathrm{cov}}_t +
r^{\mathrm{eq}}_t +
r^{\mathrm{safe}}_t +
r^{\mathrm{eff}}_t.
\end{equation}

The coverage completion term rewards step-wise improvement in the global coverage score and encourages tooth-level and global coverage targets:
\begin{equation}
r^{\mathrm{cov}}_t =
\lambda_{\mathrm{cov}}
\Delta \bar{c}_t
-
\lambda_{\mathrm{stag}}
\mathbb{I}
\left[
\Delta \bar{c}_t \leq 0
\right]
+
\lambda_{\mathrm{tooth}} b^{\mathrm{tooth}}_t
+
\lambda_{\mathrm{global}} b^{\mathrm{global}}_t,
\quad
\Delta \bar{c}_t =
\bar{c}_t - \bar{c}_{t-1}.
\end{equation}
The stagnation penalty discourages actions that do not improve the accumulated reconstruction. The tooth-level bonus $b^{\mathrm{tooth}}_t$ is a non-repeated milestone bonus computed from the combined coverage of each tooth and its adjacent gum region. The global bonus $b^{\mathrm{global}}_t$ is awarded when the global coverage score first reaches its target threshold.


To encourage balanced reconstruction across individual teeth, local tooth regions, and side surfaces,
we combine tooth-focused coverage progress, active-frontier progress, local-region completion,
undercovered-tooth departure penalties, and side-coverage rewards:
\begin{equation}
r^{\mathrm{eq}}_t =
r^{\mathrm{focus}}_t
+
r^{\mathrm{frontier}}_t
+
r^{\mathrm{region}}_t
+
r^{\mathrm{depart}}_t
+
r^{\mathrm{side}}_t .
\end{equation}
The tooth-focused term rewards improvement of a per-tooth coverage score that emphasizes
under-covered teeth:
\begin{equation}
r^{\mathrm{focus}}_t =
\lambda_{\mathrm{focus}}
\left[
S_t - S_{t-1}
\right]_+, \quad S_t = \alpha L_t + (1-\alpha) \frac{1}{K}
\sum_{k=1}^{K} c_{k,t},
\end{equation}
where $L_t$ denotes a lower-tail mean of the per-tooth coverage distribution, and $\alpha$
balances lower-tail coverage against mean per-tooth coverage. The frontier term guides the
scanner toward insufficiently observed tooth regions, and the local-region term encourages
refinement of poorly covered tooth regions. The departure penalty discourages leaving a tooth
before sufficient coverage is achieved. The side-coverage term encourages balanced buccal and
lingual acquisition.

The safety term discourages invalid scanner views and workspace violations:
\begin{equation}
r^{\mathrm{safe}}_t =
r^{\mathrm{near}}_t
+
r^{\mathrm{out}}_t .
\end{equation}
Here, $r^{\mathrm{near}}_t$ penalizes scanner poses that are too far from the nearest tooth surface, and $r^{\mathrm{out}}_t$ penalizes scanpoints outside the valid scanning region.

The efficiency term discourages unnecessary motion and long scan trajectories:
\begin{equation}
r^{\mathrm{eff}}_t =
r^{\mathrm{motion}}_t
+
r^{\mathrm{step}}_t .
\end{equation}
Here, $r^{\mathrm{motion}}_t$ penalizes excessive scanner-camera displacement between consecutive steps, and $r^{\mathrm{step}}_t$ penalizes long episodes.

\paragraph{Progressive training scheme.}
Directly enforcing high global and per-tooth coverage from the beginning can make learning sparse and unstable. We therefore use a progressive curriculum that gradually increases the coverage requirements during training.

At curriculum stage $s$, success is latched when the global coverage $\bar{c}_t$, total tooth coverage $c^{\text{teeth}}_t$, and minimum per-tooth coverage $\min_k c_{k,t}$ satisfy the stage-specific thresholds:
\begin{equation}
\bar{c}_t \geq \tau^g_s,
\quad
c^{\text{teeth}}_t \geq \tau^t_s,
\quad
\min_k c_{k,t} \geq \tau^{\min}_s .
\end{equation}
In early stages, the minimum per-tooth constraint is disabled to encourage broad exploration before enforcing strict tooth-wise balance. After the success condition is latched, the episode continues for a fixed refinement period before termination. The policy advances to the next stage when the rolling success rate over the stage-specific window of $N_s$ episodes exceeds the target rate $\rho_s$, with all active thresholds increasing progressively across stages.

\subsection{Policy Network and PPO Optimization}
\label{sec:ppo}

The actor and critic use separate two-branch networks with identical architecture and no shared parameters. The volumetric branch encodes the geometric memory $\boldsymbol{\sigma}_t$ using strided 3D convolutional layers followed by adaptive pooling and a multilayer perceptron. The proprioceptive branch encodes $\mathbf{o}^{\mathrm{prop}}_t$ using a separate multilayer perceptron. The two embeddings are concatenated and passed through a fusion network to produce either the policy output or the value estimate. This architecture separates geometric scan-memory processing from robot-state conditioning, while allowing the final decision to depend on both accumulated scan history and the current robot configuration.
We train the policy $\pi_\theta$ using Proximal Policy Optimization (PPO)~\cite{schulman2017proximal} with the advantage estimate $\hat{A}_t$ as
\begin{equation}
L^{\mathrm{CLIP}}(\theta)
=
\hat{\mathbb{E}}_t \left[ \min \left( \rho_t(\theta)\hat{A}_t,
\mathrm{clip} \left( \rho_t(\theta), 1-\epsilon, 1+\epsilon
\right) \hat{A}_t \right) \right],\quad \rho_t(\theta) = \frac{ \pi_\theta(\mathbf{a}_t \mid \mathbf{o}_t) }{ \pi_{\theta_{\mathrm{old}}}(\mathbf{a}_t \mid \mathbf{o}_t) }.
\end{equation}

\section{Experiments}
\label{sec:experiments}



\subsection{Experimental Setup}
\label{sec:exp_setup}

\paragraph{Implementation detail.} 
The policy uses a tri-state voxel input of shape $3 \times 51 \times 55 \times 30$ with actor/critic hidden layers $[256, 128]$ and ELU activations. Training uses PPO with learning rate $3 \times 10^{-4}$ and adaptive scheduling, running 32 parallel environments on 4 NVIDIA RTX 3090 GPUs for approximately 14 hours. Reward weights are provided in the supplementary material.

\paragraph{Task and dataset.}
RobOralScan is evaluated as a closed-loop robotic intraoral scanning task, where a robot-arm-mounted scanner acquires local depth observations inside a bounded oral workspace to reconstruct tooth and adjacent-gum surfaces within a fixed control horizon. For quantitative training and evaluation, we build a synthetic scanning environment from the Teeth3DS dataset~\cite{ben2022teeth3ds,ben20233dteethseg},
which provides 3D intraoral scan models with per-tooth segmentation labels. We use 80 jaw models for training and 10 held-out models from unseen patients for evaluation. In simulation, per-tooth labels are used only for reward computation and evaluation, not as policy input. 

\paragraph{Evaluation protocol and metrics.}
All methods are evaluated with the same scanner model, workspace limits, episode horizon, and reconstruction protocol. For each rollout, we compare the accumulated reconstruction $P$ with the annotated reference surface $S$. We report Chamfer distance (CD), average final coverage (Avg. Cov.), normalized coverage AUC, lower-tail per-tooth coverage, and tooth-level success. Avg. Cov. denotes the evaluator coverage averaged over tooth and adjacent-gum regions, where coverage is the fraction of reference-surface points observed within a distance threshold $\tau_{\mathrm{cov}}$. Normalized coverage AUC measures coverage accumulation over the fixed scanning horizon. We define tooth-level success by $q_T^{0.1} \geq 0.85$, where $q_T^{0.1}$ is the 10th percentile of final per-tooth coverage as $q_T^{0.1} = \mathrm{Quantile}_{0.1} \left( \{c_{k,T}\}_{k=1}^{K} \right)$. Ground-truth tooth annotations are not provided to the policy and are used only by the reward and evaluation modules.
\begin{table*}[t]
\centering
\caption{
Quantitative comparison of intraoral scanning performance. 
Rule-based and anatomy-guided methods use hand-crafted or privileged priors during planning.
Surface-Frontier NBV uses recorded end-of-episode cumulative reconstruction evaluation.
}
\label{tab:main_comparison}
\resizebox{1\linewidth}{!}{%
\begin{tabular}{llccccc}
\toprule
Category
& Method
& CD ($L_2$) $\downarrow$
& Avg. Cov. (\%) $\uparrow$
& $q_T^{0.1}$ (\%) $\uparrow$
& AUC (Norm.) $\uparrow$
& Success $\uparrow$ \\
\midrule

Scripted / Anatomy Prior
& Anatomy-Guided Dense Sweep~\cite{feng2021accuracy}
& \textbf{0.00784}
& 47.28
& \underline{3.79}
& 0.3754
& 0 / 10 \\

Classical NBV
& Volumetric Information-Gain NBV~\cite{delmerico2018comparison}
& 0.02830
& 21.73
& 0.00
& 0.1949
& 0 / 10 \\

Classical NBV
& Surface-Frontier NBV~\cite{kriegel2015efficient}
& 0.02166
& 21.69
& 0.00
& 0.1974
& 0 / 10 \\

Learning Baseline
& Coverage Completion Only
& 0.01146
& \underline{71.89}
& 2.61
& \underline{0.5183}
& 0 / 10 \\

Ours
& Full Reward (Ours)
& \underline{0.00838}
& \textbf{92.58}
& \textbf{88.45}
& \textbf{0.6674}
& \textbf{8 / 10} \\

\bottomrule
\end{tabular}%
}
\end{table*}

\subsection{Comparison with Representative Scanning Strategies}
\label{sec:main_comparison}

Table~\ref{tab:main_comparison} compares RobOralScan with representative scanning and active reconstruction strategies adapted to the same robotic intraoral scanning interface. The Anatomy-Guided Dense Sweep is a scripted anatomy-prior baseline motivated by intraoral scan-path analysis~\cite{feng2021accuracy};
it follows an approximate dental-arch trajectory without reconstruction feedback. The Volumetric Information-Gain NBV baseline follows classical NBV planning based on expected information gain~\cite{delmerico2018comparison},
whereas the Surface-Frontier NBV baseline targets frontier regions of the incomplete surface reconstruction~\cite{kriegel2015efficient}. Coverage Completion Only is an internal learned baseline that uses the same closed-loop scanner control interface as RobOralScan but optimizes only global coverage completion. Full Reward denotes RobOralScan with all proposed reward terms.
All methods are evaluated under the same scanner model, workspace limits, episode horizon, and reference-surface evaluation protocol.

The comparison reveals two main limitations of applying generic scanning strategies to robotic intraoral scanning. First, scripted and classical NBV methods fail to ensure tooth-wise completion. Although the Anatomy-Guided Dense Sweep achieves a low CD by accurately reconstructing some locally observed regions, it covers only 47.28\% of the surface with $q_T^{0.1}=3.79\%$ and no successful episodes. Volumetric Information-Gain NBV and Surface-Frontier NBV perform even worse, reaching only about 21--22\% final coverage with $q_T^{0.1}=0$. This suggests that plausible scan paths and generic geometric exploration objectives are insufficient under the short field of view, working-distance constraints, and confined oral workspace.

Second, global coverage optimization alone does not guarantee balanced tooth-level reconstruction. Coverage Completion Only improves average coverage to 71.89\%, showing the benefit of learned closed-loop control, but its $q_T^{0.1}$ remains only 2.61\% with zero successful episodes. Thus, the policy can increase overall coverage while leaving difficult teeth under-scanned. In contrast, RobOralScan with the full reward reaches 92.58\% average coverage, 88.45\% lower-tail tooth coverage, and 8/10 successful episodes. These results
support our formulation of intraoral scanning as a tooth-aware active scanning problem rather than a generic coverage maximization task.

\begin{figure}[t]
    \centering
    \includegraphics[width=\columnwidth]
    {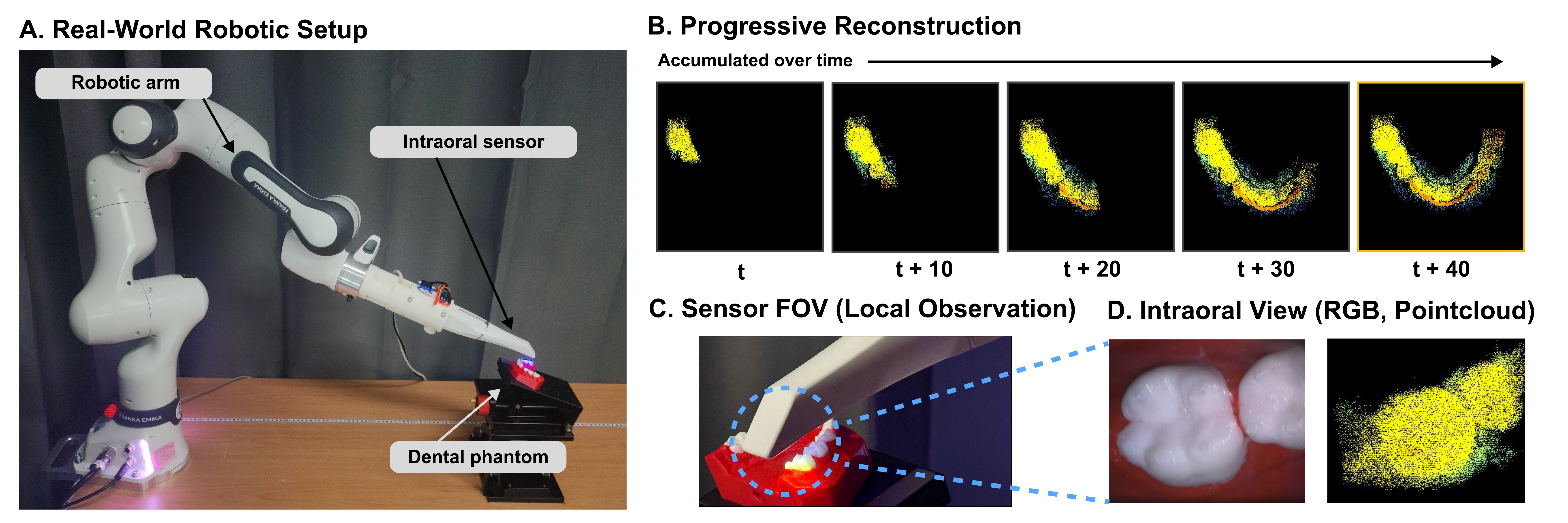}
    \caption{
    Zero-shot real-world deployment of RobOralScan on a Franka Research 3 (FR3) arm
    equipped with a Huvitz Lilivis SCAN intraoral scanner. The policy produces
    stable scanner motions and progressively reconstructs tooth and adjacent-gum
    surfaces without real-world fine-tuning.
    }
    \label{fig:real_world_rollout}
\end{figure}

\subsection{Real-world Hardware Validation}
\label{sec:real_world_validation}

We further validate RobOralScan on a real robotic intraoral scanning setup using a Franka Research 3 (FR3) arm and a Huvitz Lilivis SCAN intraoral scanner~\cite{franka_research_3,huvitz_lilivis_scan}. The policy is deployed zero-shot, without real-world fine-tuning or additional reward adaptation. As shown in Fig.~\ref{fig:real_world_rollout}, RobOralScan maintains a stable scan trajectory around the target jaw model and progressively acquires tooth and adjacent-gum surfaces despite real sensor noise, calibration error, and robot execution latency. This result suggests that the learned tooth-aware scanning
strategy transfers beyond the synthetic evaluation environment and can be executed on a physical robotic intraoral scanning platform. Additional rollout examples are provided in the supplementary video.

\subsection{Ablation Study}
\label{sec:component_ablation}

Table~\ref{tab:ablation} reports observation-space and reward ablations.

\textbf{Observation representation.}
Replacing the tri-state memory with binary occupancy drops average coverage from 92.58\% to 86.87\%, $q_T^{0.1}$ from 88.45\% to 74.50\%, and success from 8/10 to 5/10, confirming that distinguishing \textit{free} from \textit{unknown} space provides meaningful guidance for directing the scanner toward genuinely unexplored regions.

\textbf{Reward design.}
Average coverage alone is not a reliable indicator of successful intraoral scanning: the Coverage Completion Only baseline reaches 71.89\% average coverage but $q_T^{0.1}=2.61\%$ with zero successes, showing that global learning can improve easily visible regions while leaving individual teeth incomplete. Among the leave-one-out variants, removing scan safety causes the largest degradation, removing tooth-wise equity reduces success from 8/10 to 5/10 while preserving high average coverage, and removing motion efficiency lowers AUC from 0.6674 to 0.6011 without affecting final coverage. Together, the reward groups balance global progress, per-tooth completeness, safe view maintenance, and scan efficiency.

\begin{table}[t]
\centering
\scriptsize
\renewcommand{\arraystretch}{0.88}
\caption{Ablation study including geometric memory
encoding (Binary Occupancy vs.\ tri-state) and
leave-one-out reward design variants. \textbf{Bold} and \underline{underline} are the best and second best, respectively.
}
\label{tab:ablation}
\resizebox{\columnwidth}{!}{%
\begin{tabular}{lccccc}
\toprule
Variant
& CD ($L_2$) $\downarrow$
& Avg. Cov. (\%) $\uparrow$
& $q_T^{0.1}$ (\%) $\uparrow$
& AUC (Norm.) $\uparrow$
& Success $\uparrow$ \\
\midrule
Binary Occupancy
& 0.00856 & 86.87 & 74.50 & 0.6309 & 5 / 10 \\
Ours
& \underline{0.00838} & \underline{92.58}
& \textbf{88.45} & \textbf{0.6674} & \textbf{8 / 10} \\
\midrule
Coverage Completion Only
& 0.01146 & 71.89 & 2.61 & 0.5183 & 0 / 10 \\
w/o Tooth-wise Equity
& 0.00852 & \textbf{92.69} & 80.89 & 0.5901 & 5 / 10 \\
w/o Scan Safety
& 0.00953 & 79.59 & 59.31 & 0.5706 & 1 / 10 \\
w/o Motion Efficiency
& \textbf{0.00824} & 91.28
& \underline{88.25} & \underline{0.6011} & \textbf{8 / 10} \\
\bottomrule
\end{tabular}%
}
\end{table}





\section{Limitations}
\label{sec:limitations}


A limitation of RobOralScan is that the current hardware demonstration was not conducted inside a human oral cavity. 
Although the robot-scanner system supports repeatable scanner motion, in-mouth deployment may introduce additional sources of error, including restricted access, soft-tissue interaction, patient motion, and visibility changes. 
In particular, discrepancies between the policy-predicted relative action and the robot's actual executed motion can make the estimated scanner pose inaccurate. 
Such pose uncertainty may affect the accumulated geometric memory, since scan frames are integrated based on the assumed camera pose. 
Future work will therefore evaluate RobOralScan in more realistic in-mouth or human-subject settings and analyze how action-execution errors propagate to pose estimation and reconstruction quality. 
We will also investigate closed-loop pose correction using real scanner feedback, improved hand-eye calibration, and online registration to make the system more robust for clinical deployment.

\section{Conclusion}
\label{sec:conclusion}
We proposed RobOralScan, to the best of our knowledge, the first RL-based pipeline for robotic automatic intraoral scanning. The main contribution is a tooth-aware closed-loop scan-control formulation that couples accumulated geometric memory with robot state, enabling the policy to reason over scan history, insufficiently observed tooth surfaces, and feasible scanner motion inside the oral workspace. RobOralScan introduces a tri-state geometric memory representation for partial scan accumulation and a tooth-wise coverage learning objective that jointly encourages reconstruction progress, balanced per-tooth completion, scan safety, and motion efficiency. Together, these components turn intraoral scanning into an adaptive 6-DoF scanner-control problem under short-FOV, working-distance, and confined-workspace constraints. Experiments show that RobOralScan improves tooth-level reconstruction completeness and demonstrates zero-shot hardware feasibility on a physical robot-scanner system.


\clearpage


\bibliography{paper/references}

\end{document}


\maketitle

\appendix

\providecommand{\fillin}[1]{\textcolor{red}{[fill in #1]}}

\section*{Overview}

This supplementary material provides implementation details and additional experimental analyses for RobOralScan.
It complements the main paper with reproducibility details, ablation diagnostics, and zero-shot hardware deployment information.

The supplement is organized as follows.
Appendix~\ref{app:reward} gives the full reward function and progressive curriculum schedule.
Appendix~\ref{app:env} describes the simulation environment, scanner model, observation/action interface, simulated perception-to-action timing, policy architecture, and PPO training configuration.
Appendix~\ref{app:metrics} defines the evaluation metrics used for reconstruction quality, coverage, coverage AUC, and tooth-level success.
Appendix~\ref{app:baselines} specifies the implementation details of the compared baselines.
Appendix~\ref{app:extra_results} reports curriculum ablation results, executed steps, and terminal-condition diagnostics.
Appendix~\ref{app:hardware} describes the zero-shot hardware deployment setup and rollout protocol.


\section{Reward Function and Progressive Curriculum}
\label{app:reward}

\subsection{Total Reward}

We use the coverage notation defined in the main paper.
In particular, $\bar{c}_t$ denotes the global coverage score, and $c_{k,t}$
denotes the per-tooth coverage used by the tooth-wise training signals.
For the reward terms, the per-tooth milestone and tooth-wise equity signals use
the implementation convention described in the main paper: coverage of tooth
$k$ is computed together with its associated adjacent gingival region.

The total reward follows the grouping used in the main paper:
\begin{equation}
r_t
=
r_t^{\mathrm{cov}}
+
r_t^{\mathrm{eq}}
+
r_t^{\mathrm{safe}}
+
r_t^{\mathrm{eff}}.
\end{equation}
The coverage term measures global reconstruction progress through $\bar{c}_t$,
whereas the tooth-wise equity term uses per-tooth statistics computed from
$\{c_{k,t}\}_{k=1}^{K}$.

\subsection{Coverage Completion Reward}

The coverage completion reward encourages the agent to discover and scan new tooth surfaces. It combines immediate progress tracking with milestone-based bonuses:
\begin{equation}
r_t^{\mathrm{cov}}
=
\lambda_{\mathrm{cov}}
\left(
\Delta \bar{c}_t
-
\beta_{\mathrm{stag}}
\mathbf{1}
[
\Delta \bar{c}_t \leq 0
]
\right)
+
\lambda_{\mathrm{tooth}} b_t^{\mathrm{tooth}}
+
\lambda_{\mathrm{global}} b_t^{\mathrm{global}},
\end{equation}
where $\Delta \bar{c}_t = \bar{c}_t-\bar{c}_{t-1}$ represents the change in global surface-level coverage. If no new coverage is achieved ($\Delta \bar{c}_t \leq 0$), a stagnation penalty $\beta_{\mathrm{stag}}$ is applied to prevent the scanner from idling in already-scanned regions.

\paragraph{Per-Tooth Milestone Bonus ($b_t^{\mathrm{tooth}}$)}
To alleviate the challenge of sparse rewards in large environment spaces, we provide intermediate milestone bonuses for individual teeth:
\begin{equation}
b_t^{\mathrm{tooth}}
=
\sum_{k=1}^{K}
\left(
\beta_{\mathrm{hit}}h_{k,t}
+
\beta_{\mathrm{mile}}
\sum_{m\in\mathcal{M}}
q_{k,t}^{m}
\right),
\end{equation}
where $h_{k,t} \in \{0, 1\}$ indicates whether tooth $k$ is hit for the first time at step $t$, and $q_{k,t}^{m} \in \{0, 1\}$ indicates whether tooth $k$ reaches a specific coverage milestone $m \in \mathcal{M}$ for the first time, using the milestone set listed in Table~\ref{tab:reward_params}.

\paragraph{Global Completion Bonus ($b_t^{\mathrm{global}}$)}
The global completion bonus $b_t^{\mathrm{global}}$ is awarded as a one-time sparse reward when the global coverage $\bar{c}_t$ first crosses the threshold required by the current curriculum stage.

\subsection{Tooth-Wise Equity Reward}

To prevent the agent from over-scanning easily accessible areas (e.g., occlusal surfaces) while neglecting difficult ones, we introduce a tooth-wise equity reward. This term comprises five components targeting distinct intra-oral scanning behaviors:
\begin{equation}
r_t^{\mathrm{eq}}
=
r_t^{\mathrm{focus}}
+
r_t^{\mathrm{frontier}}
+
r_t^{\mathrm{region}}
+
r_t^{\mathrm{depart}}
+
r_t^{\mathrm{side}}.
\end{equation}

\paragraph{Focus Reward ($r_t^{\mathrm{focus}}$)} 
This term prioritizes the least-scanned teeth by evaluating the step-by-step progress of a focused score $S_t$:
\begin{equation}
r_t^{\mathrm{focus}}
=
\lambda_{\mathrm{focus}} \left[ S_t-S_{t-1} \right]_+,
\qquad
S_t = \alpha L_t + (1-\alpha)m_t,
\end{equation}
where $m_t = \frac{1}{K} \sum_{k=1}^{K} c_{k,t}$ is the mean coverage across all $K$ teeth, and $L_t$ is the mean coverage of the lowest-covered tail fraction ($q_{\mathrm{tail}}$) of teeth. The lower-tail set size is bounded by $K_{\mathrm{tail}} = \max (1, \lceil q_{\mathrm{tail}}K \rceil)$ after sorting the per-tooth coverages ($c_{(1),t} \le \dots \le c_{(K),t}$).

\paragraph{Frontier Reward ($r_t^{\mathrm{frontier}}$)} 
To promote continuous exploration toward unscanned boundaries, we reward reductions in the distance to the active-tooth frontier, denoted by $d_t^{\mathrm{frontier}}$:
\begin{equation}
r_t^{\mathrm{frontier}}
=
\lambda_{\mathrm{frontier}}
\min
\left(
\frac{ \left[ d_{t-1}^{\mathrm{frontier}} - d_t^{\mathrm{frontier}} \right]_+ }{s_{\mathrm{frontier}}},
r_{\mathrm{frontier}}^{\max}
\right).
\end{equation}

\paragraph{Local-Region Completion ($r_t^{\mathrm{region}}$)} 
This term shifts focus from global structures to the fine-grained details within the currently active tooth. Let $u_{j,t}$ be the coverage of local region $j \in \{1, \dots, J\}$ on the active tooth. The regional score $U_t$ is defined as:
\begin{equation}
r_t^{\mathrm{region}}
=
\lambda_{\mathrm{region}} \left[ U_t-U_{t-1} \right]_+,
\qquad
U_t = w_{\mathrm{low}} Q_{q_{\mathrm{region}}} (\{u_{j,t}\}_{j=1}^{J}) + w_{\mathrm{mean}} \frac{1}{J} \sum_{j=1}^{J} u_{j,t},
\end{equation}
where $Q_{q_{\mathrm{region}}}(\cdot)$ computes the specified percentile of the local regional coverages to penalize sub-tooth blind spots.

\paragraph{Departure Penalty ($r_t^{\mathrm{depart}}$)} 
To eliminate inefficient "hit-and-run" behaviors, the agent is penalized if it prematurely moves away ($d_{k,t} \ge d_{\mathrm{leave}}$) from an approached tooth ($A_{k,t}=1$) before achieving a sufficient coverage threshold ($\tau_{\mathrm{depart}}$):
\begin{equation}
r_t^{\mathrm{depart}}
=
-\lambda_{\mathrm{depart}}\beta_{\mathrm{depart}}
\sum_{k=1}^{K}
\mathbf{1}
\left[ A_{k,t}=1 \land c_{k,t}<\tau_{\mathrm{depart}} \land d_{k,t}\ge d_{\mathrm{leave}} \land D_{k,t-1}=0 \right],
\end{equation}
where the tracking variable $D_{k,t-1}=0$ ensures that this penalty is issued at most once per tooth.

\paragraph{Side Coverage Reward ($r_t^{\mathrm{side}}$)} 
To target notoriously hard-to-reach areas, we explicitly reward improvements in the worst-covered buccal and lingual sides across the entire dental arch:
\begin{equation}
r_t^{\mathrm{side}}
=
\lambda_{\mathrm{buccal}} \left[ \mu_t^{\mathrm{buccal}} - \mu_{t-1}^{\mathrm{buccal}} \right]_+
+
\lambda_{\mathrm{lingual}} \left[ \mu_t^{\mathrm{lingual}} - \mu_{t-1}^{\mathrm{lingual}} \right]_+,
\end{equation}
where the minimum side coverages are tracked via $\mu_t^{\mathrm{buccal}} = \min_{k} c_{k,t}^{\mathrm{buccal}}$ and $\mu_t^{\mathrm{lingual}} = \min_{k} c_{k,t}^{\mathrm{lingual}}$.

\subsection{Safety and Efficiency Rewards}

The safety reward penalizes invalid scanner poses, while the efficiency reward
penalizes excessive motion and long episodes. The safety term is decomposed as
\begin{equation}
r_t^{\mathrm{safe}}
=
r_t^{\mathrm{near}}
+
r_t^{\mathrm{out}}.
\end{equation}
The efficiency term $r_t^{\mathrm{eff}}$ is defined below.

\paragraph{Working-Distance Penalty ($r_t^{\mathrm{near}}$)}
To keep the scanner within the valid target-distance region, the agent is
penalized as the nearest-tooth distance $d_t$ increases beyond the preferred
activation distance $d_{\mathrm{near}}^{\mathrm{start}}$:
\begin{equation}
r_t^{\mathrm{near}}
=
-\lambda_{\mathrm{near}}\beta_{\mathrm{near}}
\operatorname{clip}
\left(
\frac{
d_t-d_{\mathrm{near}}^{\mathrm{start}}
}{
d_{\mathrm{near}}^{\mathrm{full}}-d_{\mathrm{near}}^{\mathrm{start}}
},
0,
1
\right).
\end{equation}
Here, $d_{\mathrm{near}}^{\mathrm{full}}$ denotes the distance at which the
full working-distance penalty is applied. The penalty therefore increases as
the scanner moves farther away from the tooth surface beyond the preferred
working-distance region.

\paragraph{Workspace Boundary Penalty ($r_t^{\mathrm{out}}$)}
To restrict the scanner within a valid operational envelope, a steep penalty is imposed if the position of the scanner-camera scanpoint $x_t$ deviates beyond an acceptable radius $d_{\mathrm{out}}$ from the dental target center $x_{\mathrm{center}}^{\mathrm{teeth}}$:
\begin{equation}
r_t^{\mathrm{out}}
=
-\lambda_{\mathrm{out}}\beta_{\mathrm{out}}
\mathbf{1}
\left[
\left\| x_t-x_{\mathrm{center}}^{\mathrm{teeth}} \right\|_2 > d_{\mathrm{out}}
\right].
\end{equation}

\paragraph{Kinematic Efficiency Penalty ($r_t^{\mathrm{eff}}$)}
To encourage smooth, non-redundant actions and rapid scanning completion, we penalize both spatial displacement and the temporal progression of the episode:
\begin{equation}
r_t^{\mathrm{eff}}
=
-\lambda_{\mathrm{motion}} \left\| x_t-x_{t-1} \right\|_2 - \lambda_{\mathrm{step}} \frac{n_t}{T},
\end{equation}
where $\|x_t-x_{t-1}\|_2$ is the Euclidean distance traveled in the current step, $n_t$ is the current episode step, and $T$ is the maximum allowable episode length.

\subsection{Reward Parameters}
\label{app:reward_weights}

All reported experiments use the fixed reward parameters listed in
Table~\ref{tab:reward_params}.

\begin{table}[!tbp]
\centering
\caption{Reward parameters used in the reported experiments.}
\label{tab:reward_params}
\small
\setlength{\tabcolsep}{4pt}
\renewcommand{\arraystretch}{1.08}
\begin{tabular}{p{0.15\linewidth}p{0.47\linewidth}p{0.30\linewidth}}
\toprule
Group & Parameter & Value \\
\midrule
Coverage
& Coverage progress coefficient, $\lambda_{\mathrm{cov}}$
& $40.0$ \\
Coverage
& Stagnation offset, $\beta_{\mathrm{stag}}$
& $0.05$ \\
Coverage
& Per-tooth bonus coefficient, $\lambda_{\mathrm{tooth}}$
& $1.0$ \\
Coverage
& First-hit / milestone bonus, $\beta_{\mathrm{hit}}/\beta_{\mathrm{mile}}$
& $0.05 / 0.25$ \\
Coverage
& Per-tooth milestone set, $\mathcal{M}$
& $\{0.2,0.4,0.6,0.8,$ \newline $0.9,0.95,0.98,1.0\}$ \\
Coverage
& Global completion coefficient, $\lambda_{\mathrm{global}}$
& $10.0$ \\
\midrule
Equity
& Tooth-focused coefficient, $\lambda_{\mathrm{focus}}$
& $12.0$ \\
Equity
& Tail fraction / lower-tail score weight / mean score weight
& $0.10 / 0.65 / 0.35$ \\
Equity
& Frontier coefficient / distance scale / reward cap
& $0.5 / 0.01 / 1.0$ \\
Equity
& Regional coefficient / percentile / low-percentile--mean weights
& $2.0 / 0.20 / 0.75$--$0.25$ \\
Equity
& Minimum regional samples
& $8$ \\
Equity
& Departure coefficient / raw penalty magnitude
& $1.0 / 0.25$ \\
Equity
& Undercoverage / approach / leave thresholds
& $0.70 / 0.022\,\mathrm{m} / 0.035\,\mathrm{m}$ \\
Equity
& Buccal / lingual side-progress coefficients
& $8.0 / 8.0$ \\
\midrule
Safety
& Nearest-distance coefficient / raw penalty magnitude
& $1.5 / 2.0$ \\
Safety
& Working-distance penalty start / full-penalty distance
& $0.022\,\mathrm{m} / 0.050\,\mathrm{m}$ \\
Safety
& Scanpoint-out coefficient / raw penalty magnitude / threshold
& $1.0 / 50.0 / 0.08\,\mathrm{m}$ \\
\midrule
Efficiency
& Motion / step coefficients
& $0.1 / 0.4$ \\
\bottomrule
\end{tabular}
\end{table}

\subsection{Progressive Curriculum}

Table~\ref{tab:curriculum} gives the full curriculum schedule.
At stage $s$, success requires the listed global coverage, tooth-surface coverage,
and minimum per-tooth coverage thresholds.
The policy advances to the next stage when the rolling success rate over $N_s$
episodes reaches $\rho_s$.
Tail-refinement stages require the policy to continue scanning for the listed
number of additional steps after first satisfying the coverage condition.

\begin{table}[!tbp]
\centering
\caption{
Progressive curriculum schedule.
Coverage thresholds are reported in percent.
A dash indicates that no minimum per-tooth threshold is applied at that stage.
}
\label{tab:curriculum}
\small
\setlength{\tabcolsep}{5pt}
\renewcommand{\arraystretch}{1.05}
\begin{tabular}{ccccccc}
\toprule
Stage $s$
& $\tau_s^g$ (\%)
& $\tau_s^t$ (\%)
& $\tau_s^{\min}$ (\%)
& $\rho_s$
& $N_s$
& Refinement steps \\
\midrule
0  & 75.0 & 75.0 & --   & 0.28 & 120 & 0  \\
1  & 78.0 & 80.0 & 30.0 & 0.36 & 150 & 0  \\
2  & 84.0 & 86.0 & 45.0 & 0.44 & 180 & 0  \\
3  & 88.0 & 90.0 & 55.0 & 0.52 & 220 & 0  \\
4  & 90.0 & 92.0 & 63.0 & 0.54 & 260 & 15 \\
5  & 90.0 & 92.0 & 68.0 & 0.60 & 300 & 15 \\
6  & 91.0 & 93.0 & 72.0 & 0.71 & 380 & 15 \\
7  & 91.5 & 93.5 & 76.0 & 0.75 & 430 & 15 \\
8  & 92.0 & 94.0 & 80.0 & 0.79 & 500 & 20 \\
9  & 92.5 & 94.5 & 84.0 & 0.83 & 580 & 20 \\
10 & 93.0 & 95.0 & 88.0 & 0.86 & 660 & 20 \\
\bottomrule
\end{tabular}
\end{table}

\FloatBarrier


\section{Simulation Environment, Control Interface, and Training Details}
\label{app:env}

\subsection{Observation, Action, and Control Interface}

The policy observes a tri-state geometric memory and a proprioceptive robot-state vector.
Table~\ref{tab:obs_action_control} summarizes the scanner input, action space, and control interface.

\begin{table}[H]
\centering
\caption{Observation, action, and control interface.}
\label{tab:obs_action_control}
\small
\setlength{\tabcolsep}{4pt}
\renewcommand{\arraystretch}{1.08}
\begin{tabular}{@{}p{0.25\linewidth}p{0.69\linewidth}@{}}
\toprule
Item & Specification \\
\midrule
Depth image 
& $120 \times 144$, depth range $0.001$--$0.022$ m \\

Geometric observation 
& One-hot tri-state voxel memory over unknown, free, and occupied states \\

Voxel grid 
& $51 \times 55 \times 30$ grid with $0.002$ m isotropic voxels over a jaw-centered scanning workspace \\

Proprioception 
& 21-D vector: joint positions (7), joint velocities (7), scanpoint position (3), scanpoint quaternion (4) \\

Action 
& Relative 6-DoF scanner motion in the scanner TCP frame \\

Translation scale 
& $(0.01,0.01,0.01)$ m \\

Rotation scale 
& $(0.04,0.04,0.04)$ rad \\

Control interface 
& Damped least-squares differential IK with damping coefficient $\lambda=0.01$ \\
\bottomrule
\end{tabular}
\end{table}

\paragraph{Simulated perception-to-action timing.}
We profile the simulated perception-to-action path using the scanner input configuration in Table~\ref{tab:obs_action_control}.
Tri-state preprocessing converts each depth frame into the policy's voxel memory through depth-to-world point-cloud conversion, voxel downsampling, occupied-voxel indexing, and free-space ray-casting.
Coverage-tracker update denotes the KDTree-based reward/evaluation coverage computation used in simulation.
These measurements are simulation-side timing values and do not include hardware communication, scanner-driver latency, or robot execution latency.

\begin{table}[H]
\centering
\caption{Simulated perception-to-action timing and update rates.}
\label{tab:perception_action_timing}
\small
\setlength{\tabcolsep}{4pt}
\renewcommand{\arraystretch}{1.08}
\begin{tabular}{@{}p{0.33\linewidth}p{0.29\linewidth}cc@{}}
\toprule
Component & Measurement & Value / Mean & P95 \\
\midrule
Robot control loop 
& Update rate 
& 100 Hz 
& -- \\

Policy/scanner update loop 
& Update rate 
& 20 Hz 
& -- \\

Policy forward 
& GPU, batch size 1 
& 0.50 ms 
& 0.53 ms \\

Policy forward 
& CPU 
& 2.59 ms 
& 2.76 ms \\

Tri-state preprocessing 
& No coverage-tracker update 
& 22.25 ms 
& 23.47 ms \\

Coverage-tracker update 
& KDTree + visibility gate 
& 4.73 ms 
& 4.88 ms \\

Full simulated step 
& Preprocess + tracker + policy 
& 27.5 ms 
& $\approx 29$ ms \\
\bottomrule
\end{tabular}
\end{table}

The policy network itself is not the runtime bottleneck.
Most latency comes from constructing the tri-state voxel input, especially free-space ray-casting.
The full simulated preprocessing--tracking--policy path remains below the 50 ms period of the 20 Hz policy/scanner update loop.

\subsection{Policy Architecture and PPO Training}

All learning-based policies use the same observation space, action space, policy architecture, PPO optimizer, scanner model, and simulator.
For reward ablations, only the reward terms are changed.

\begin{table}[H]
\centering
\caption{PPO hyperparameters for learning-based methods.}
\label{tab:ppo_hyperparams}
\small
\setlength{\tabcolsep}{4pt}
\renewcommand{\arraystretch}{1.08}
\begin{tabular}{@{}p{0.34\linewidth}p{0.56\linewidth}@{}}
\toprule
Hyperparameter & Value \\
\midrule
Algorithm 
& PPO with GAE \\

Optimizer and schedule 
& Adam, learning rate $3.0\times10^{-4}$, adaptive learning-rate schedule \\

Rollout length 
& 16 environment steps per parallel environment \\

Batching 
& 2048 transitions per PPO update; 4 minibatches of 512 transitions \\

PPO update 
& 4 epochs, clip range $\epsilon=0.2$ \\

Return estimation 
& Discount factor $\gamma=0.99$, GAE parameter $\lambda=0.95$ \\

Loss weights 
& Value loss coefficient $1.0$, entropy coefficient $0.001$ \\

Stabilization 
& Maximum gradient norm $1.0$, initial action standard deviation $1.0$ \\

Observation normalization 
& Disabled for actor and critic observations \\

Training compute 
& 4 NVIDIA RTX 3090 GPUs, approximately 14 hours \\
\bottomrule
\end{tabular}
\end{table}

The policy is a non-recurrent actor--critic network.
Actor and critic use the same architecture but have separate parameters.
Each network has a voxel branch for tri-state memory, a proprioception branch for robot state, a fusion MLP, and an output head.

\begin{table}[H]
\centering
\caption{Policy network structure.}
\label{tab:policy_network}
\small
\setlength{\tabcolsep}{4pt}
\renewcommand{\arraystretch}{1.08}
\begin{tabular}{@{}p{0.30\linewidth}p{0.60\linewidth}@{}}
\toprule
Component & Architecture \\
\midrule
Voxel input 
& One-hot tri-state grid, $3\times51\times55\times30$ \\

Voxel encoder 
& Conv3D $3{\rightarrow}16{\rightarrow}32{\rightarrow}64$ with kernel size $3$, stride $2$, padding $1$; BatchNorm3D and ReLU after each convolution \\

Voxel pooling / post-MLP 
& AdaptiveAvgPool3D to $2\times2\times2$, flatten to 512-D, then Linear $512{\rightarrow}256{\rightarrow}128$ with ELU \\

Proprioception encoder 
& 21-D input, then Linear $21{\rightarrow}128{\rightarrow}64$ with ELU \\

Fusion MLP 
& Concatenate 128-D voxel and 64-D proprioception embeddings, then Linear $192{\rightarrow}256{\rightarrow}128$ with ELU \\

Actor output 
& 6-D Gaussian action mean with state-independent scalar standard deviation initialized to $1.0$ \\

Critic output 
& Scalar value \\

Sharing / normalization 
& Separate actor and critic networks; BatchNorm3D only in the voxel encoder; no actor/critic observation normalization \\
\bottomrule
\end{tabular}
\end{table}

\FloatBarrier
\section{Evaluation Metrics}
\label{app:metrics}

Let $\bar{P}_t$ denote the accumulated reconstruction after evaluation step $t$,
and let $S$ denote the annotated reference surface. We decompose the reference
surface into tooth surfaces $S^{\mathrm{teeth}}$, tooth-adjacent gingival
surfaces $S^{\mathrm{adj\mbox{-}gum}}$, and per-tooth surfaces
$\{S_k^{\mathrm{tooth}}\}_{k=1}^{K}$.

\subsection{Chamfer Distance}

We report the L2 Chamfer distance between the final accumulated reconstruction
$\bar{P}_{T_{\mathrm{exec}}}$ and the annotated reference surface $S$:
\begin{equation}
\mathrm{CD}_{L2}(\bar{P}_{T_{\mathrm{exec}}},S)
=
\frac{1}{|\bar{P}_{T_{\mathrm{exec}}}|}
\sum_{p \in \bar{P}_{T_{\mathrm{exec}}}}
\min_{s \in S}
\|p-s\|_2
+
\frac{1}{|S|}
\sum_{s \in S}
\min_{p \in \bar{P}_{T_{\mathrm{exec}}}}
\|s-p\|_2 .
\end{equation}

\subsection{Coverage}

For any reference set $S' \subseteq S$, the coverage of $S'$ by reconstruction
$\bar{P}_t$ is
\begin{equation}
\mathrm{Cov}
\left(
\bar{P}_t,S';\tau_{\mathrm{cov}}
\right)
=
\frac{1}{|S'|}
\sum_{s \in S'}
\mathbf{1}
\left[
\min_{p \in \bar{P}_t}
\|s-p\|_2
\leq
\tau_{\mathrm{cov}}
\right].
\end{equation}

The tooth-surface and tooth-adjacent gingival coverages are
\begin{equation}
c_t^{\mathrm{teeth}}
=
\mathrm{Cov}
\left(
\bar{P}_t,
S^{\mathrm{teeth}};
\tau_{\mathrm{cov}}
\right),
\qquad
c_t^{\mathrm{adj\mbox{-}gum}}
=
\mathrm{Cov}
\left(
\bar{P}_t,
S^{\mathrm{adj\mbox{-}gum}};
\tau_{\mathrm{cov}}
\right).
\end{equation}

The average coverage score is
\begin{equation}
\bar{c}_t
=
\frac{1}{2}
\left(
c_t^{\mathrm{teeth}}
+
c_t^{\mathrm{adj\mbox{-}gum}}
\right).
\end{equation}
The reported final average coverage is
\begin{equation}
\mathrm{AvgCov}
=
\bar{c}_{T_{\mathrm{exec}}},
\end{equation}
where $T_{\mathrm{exec}}$ denotes the final recorded evaluation step.

\subsection{Normalized Coverage AUC}

The normalized coverage AUC measures how quickly coverage accumulates over the fixed evaluation horizon $T$.
If an episode terminates before the fixed horizon, the final accumulated coverage is held constant for the remaining horizon:
\begin{equation}
\mathrm{AUC}_{\mathrm{norm}}
=
\frac{1}{T}
\sum_{t=1}^{T}
\bar{c}_{\min(t,T_{\mathrm{exec}})}.
\end{equation}
Here, $T_{\mathrm{exec}}$ denotes the final executed evaluation step.
Executed steps and terminal conditions are reported separately.

\subsection{Tooth-Level Success}

For tooth-level evaluation, we compute the final per-tooth tooth-surface coverage:
\begin{equation}
c_{k,T}^{\mathrm{eval}}
=
\mathrm{Cov}
\left(
\bar{P}_{T_{\mathrm{exec}}},
S_k^{\mathrm{tooth}};
\tau_{\mathrm{cov}}
\right).
\end{equation}
If an episode terminates before the fixed horizon, the final accumulated reconstruction is used for the remaining horizon.

The 10th-percentile per-tooth coverage is
\begin{equation}
q_T^{0.1}
=
\mathrm{Quantile}_{0.1}
\left(
\{c_{k,T}^{\mathrm{eval}}\}_{k=1}^{K}
\right).
\end{equation}

We define the reported tooth-level success indicator as
\begin{equation}
q_T^{0.1} \ge 0.85.
\end{equation}

This statistic is separate from the training-time lower-tail mean $L_t$, which is computed at every training step for the tooth-focused reward.

\section{Implementation Details of Compared Baselines}
\label{app:compared_baseline_impl}
\label{app:baselines}

This section describes how the baselines in the main comparison are
implemented in our intraoral scanning environment. The cited works in the
main table indicate the baseline family that each method represents. Since
those methods are not originally designed for robot-constrained intraoral
scanning, we implement adapted versions using the same scanner model, robot
workspace, episode length, and reconstruction pipeline as RobOralScan.

\paragraph{Anatomy-Guided Dense Sweep.}
This baseline implements a scripted full-arch scan path. It uses privileged
reference jaw geometry during planning, rather than online reconstruction
feedback. The planner first estimates the tooth order along the dental arch.
It then samples target points from occlusal, buccal, and lingual tooth regions.
For each target point, the scanner is placed at a valid working distance and
oriented toward the local tooth surface. The resulting sequence is executed as
a fixed trajectory.

Candidate poses are discarded if they violate the scanner field of view, the
valid working-distance range, the dental support region, the robot workspace,
or IK feasibility. This baseline therefore represents a strong scripted
anatomy-prior path, but it does not adapt its motion based on the partial
reconstruction accumulated during the episode.

\paragraph{Volumetric Information-Gain NBV.}
This baseline implements a classical volumetric next-best-view planner. At
each step, it reads the voxel reconstruction memory and identifies unknown
volume cells as information-gain targets. Let $\sigma_t(v)$ denote the state of
voxel $v$ in the current reconstruction memory. The unknown-volume query set is
defined as
\begin{equation}
\mathcal{V}^{\mathrm{unk}}_t
=
\{v \in \mathcal{V}_t \mid \sigma_t(v)=\mathrm{unknown}\}.
\end{equation}
The planner computes a local target from this unknown-volume distribution,
bounds the target displacement around the dental support center, and samples
192 candidate scanner views using a fixed grid of azimuth angles, elevation
angles, and radial distances.

Each candidate is evaluated by projecting the unknown-volume cells into the
candidate scanner frustum. The volumetric information-gain score is
\begin{equation}
G_{\mathrm{vol}}(a)
=
\sum_{v \in \mathcal{V}^{\mathrm{unk}}_t}
\mathbf{1}
[
v \in \mathrm{Frustum}(a)
].
\end{equation}
The planner selects the valid candidate with the largest
$G_{\mathrm{vol}}(a)$. Thus, this baseline explicitly searches for the view
that exposes the largest amount of still-unknown reconstruction volume. In
implementation, candidate views are additionally filtered by the same
workspace, support-region, field-of-view, working-distance, and IK checks as
RobOralScan. The planner replans at every step.

\paragraph{Surface-Frontier NBV.}
This baseline implements a surface-frontier planner. Instead of directly
scoring unknown volume cells, it operates on the reconstructed surface cache.
The planner extracts frontier seeds from already observed surface points. A
frontier seed is an observed surface point whose local neighborhood borders
unknown or low-density reconstructed regions. These seeds form the set
$\mathcal{B}^{\mathrm{front}}_t$. The planner then generates local standoff
views around the frontier seeds using the estimated surface normal.

The frontier set is defined from the surface cache, not directly from unknown
volume:
\begin{equation}
\mathcal{B}^{\mathrm{front}}_t
=
\{
b \in \mathcal{S}_t
\mid
\mathcal{N}(b)
\text{ contains unknown or low-density reconstructed regions}
\}.
\end{equation}
For a candidate view $a$, the surface-frontier score is
\begin{equation}
G_{\mathrm{frontier}}(a)
=
\sum_{b \in \mathcal{B}^{\mathrm{front}}_t}
w_t(b)
\mathbf{1}
[
b \in \mathrm{Frustum}(a)
]
\mathbf{1}
[
\mathrm{normal}(b) \text{ faces the scanner}
],
\end{equation}
where $w_t(b)$ emphasizes boundary and low-density frontier points. The planner
selects the valid candidate with the largest $G_{\mathrm{frontier}}(a)$. Thus,
this baseline follows the currently observed surface boundary, rather than
directly counting unknown volume cells.

Candidate views are filtered using the same workspace, support-region,
field-of-view, working-distance, and IK checks as RobOralScan. For the results
reported in the main table, the Surface-Frontier NBV row is evaluated from the
recorded cumulative reconstruction at the end of each episode, which is why the
main table explicitly notes its evaluation protocol.

\paragraph{Coverage Completion Only.}
This baseline is a learning-based reward ablation. It uses the same policy
architecture, observation space, action space, PPO optimizer, training
environment, scanner model, and episode length as RobOralScan. The reward keeps
the basic coverage-completion signals: global coverage progress, the
non-improving-step stagnation penalty, per-tooth completion milestones,
global-completion bonus, and the shared safety and efficiency penalties. In
compact form,
\begin{equation}
r_t^{\mathrm{CCO}}
=
r_t^{\mathrm{cov}}
+
r_t^{\mathrm{safe}}
+
r_t^{\mathrm{eff}}.
\end{equation}
Here, $r_t^{\mathrm{cov}}$ is the coverage completion reward defined in
Appendix~\ref{app:reward}, including global coverage progress, the stagnation
penalty, per-tooth milestone bonuses, and the global completion bonus.

Compared with the full RobOralScan reward, this baseline removes the
tooth-wise equity term $r_t^{\mathrm{eq}}$. Therefore, it does not include the
terms that explicitly target hard-to-complete regions: active-tooth focus
progress, frontier-progress shaping, low-percentile local-region completion,
buccal and lingual side-surface completion, and penalties for moving away from
under-covered teeth. This baseline tests whether generic coverage-completion
signals are sufficient without the proposed tooth-completion reward design.
\section{Additional Episode Statistics and Qualitative Analysis}
\label{app:extra_results}

Throughout this section, ``Exec. steps'' reports the average number of executed
steps per episode, and termination counts are reported out of 10 episodes.
``Success'' denotes episodes satisfying the final success condition,
``Horizon'' denotes episodes reaching the maximum evaluation horizon, and
``Far'' denotes episodes terminated because the scanner moved outside the valid
target-distance region used by the implementation.

\subsection{Curriculum Ablation}
\label{app:curriculum_ablation}

The main paper evaluates the effect of the tri-state memory representation and
the major reward groups. Here, we additionally ablate the progressive curriculum,
which affects the density of the success signal during early training. Without
the curriculum, the final high-coverage and tooth-wise success criteria are
enforced from the beginning, making successful episodes rare during early policy
optimization.

\begin{table}[!htbp]
\centering
\caption{
Additional ablation on curriculum learning over 10 held-out evaluation episodes.
Coverage and $q_T^{0.1}$ are reported as percentages.
}
\label{tab:curriculum_ablation}
\resizebox{\linewidth}{!}{%
\begin{tabular}{lcccccccc}
\toprule
Variant
& \makecell{CD\\$(L_2)\downarrow$}
& \makecell{Avg. Cov.\\(\%) $\uparrow$}
& \makecell{$q_T^{0.1}$\\(\%) $\uparrow$}
& \makecell{AUC\\(Norm.) $\uparrow$}
& \makecell{Exec.\\steps}
& Success
& Horizon
& Far \\
\midrule
Full Reward (Ours)
& 0.00838
& 92.58
& 88.45
& 0.6674
& 25.7
& 8 / 10
& 0 / 10
& 2 / 10 \\

w/o Curriculum
& 0.02343
& 29.00
& 2.18
& 0.2372
& 87.7
& 0 / 10
& 1 / 10
& 9 / 10 \\
\bottomrule
\end{tabular}%
}
\end{table}

Table~\ref{tab:curriculum_ablation} shows that directly optimizing the final
task criterion from the beginning is insufficient for this sparse-success active
scanning problem. Without curriculum learning, the policy executes longer
episodes but obtains substantially lower final coverage and almost no lower-tail
tooth coverage. The large number of far-distance terminations further suggests
unstable scanner placement rather than simple early stopping, supporting the use
of progressive coverage thresholds before enforcing the final tooth-wise success
condition.

\subsection{Episode Diagnostics for Observation and Reward Ablations}
\label{app:ablation_diagnostics}

The main paper reports aggregate reconstruction metrics for both the
observation-space ablation and the reward-design ablations. Here, we additionally
report executed steps and terminal-condition counts from the same held-out
evaluation rollouts used for the main ablation table. Binary Occupancy replaces
the tri-state memory with a binary occupied/non-occupied representation. For the
leave-one-out reward variants, w/o Tooth-wise Equity, w/o Scan Safety, and w/o
Motion Efficiency remove the corresponding reward group while keeping the
remaining training and evaluation protocol unchanged.

\begin{table}[!htbp]
\centering
\caption{
Episode-level termination diagnostics for observation and reward ablations over
10 held-out evaluation episodes. This table complements the aggregate ablation
metrics in the main paper by reporting executed steps and terminal-condition
counts.
}
\label{tab:ablation_diagnostics}
\small
\setlength{\tabcolsep}{5pt}
\renewcommand{\arraystretch}{1.08}
\begin{tabular}{@{}llcccc@{}}
\toprule
Ablation group
& Variant
& \makecell{Exec.\\steps}
& Success
& Horizon
& Far \\
\midrule
\multirow{2}{*}{\makecell[l]{Observation-space ablation}}
& Binary Occupancy
& 35.9
& 5 / 10
& 0 / 10
& 5 / 10 \\

& Full Reward (Ours)
& 25.7
& 8 / 10
& 0 / 10
& 2 / 10 \\

\midrule
\multirow{4}{*}{\makecell[l]{Reward-design ablations}}
& Coverage Completion Only
& 14.9
& 0 / 10
& 0 / 10
& 10 / 10 \\

& w/o Tooth-wise Equity
& 22.6
& 5 / 10
& 0 / 10
& 5 / 10 \\

& w/o Scan Safety
& 15.2
& 1 / 10
& 0 / 10
& 9 / 10 \\

& w/o Motion Efficiency
& 27.3
& 8 / 10
& 0 / 10
& 2 / 10 \\
\bottomrule
\end{tabular}
\end{table}

Table~\ref{tab:ablation_diagnostics} indicates that the ablated policies fail
through different termination modes. Binary Occupancy requires longer execution,
suggesting that the free/unknown distinction helps the policy identify
unexplored regions more efficiently. Coverage Completion Only and w/o Scan
Safety mostly fail by far-distance termination, highlighting the importance of
tooth-relative scanner placement and scan-safety rewards. In contrast, w/o Motion
Efficiency preserves the success rate but slightly increases execution length.

\begin{table}[!htbp]
\centering
\caption{
Representative qualitative reconstructions for reward-design variants. Each
panel shows the accumulated scan result for the corresponding policy variant.
}
\label{tab:reward_ablation_qualitative}
\scriptsize
\setlength{\tabcolsep}{2pt}
\renewcommand{\arraystretch}{1.05}
\begin{tabular}{@{}ccccc@{}}
\toprule
\makecell{Coverage\\Completion\\Only}
& \makecell{Full Reward\\(Ours)}
& \makecell{w/o\\Tooth-wise\\Equity}
& \makecell{w/o\\Scan\\Safety}
& \makecell{w/o\\Motion\\Efficiency} \\
\midrule
\includegraphics[width=0.18\linewidth]{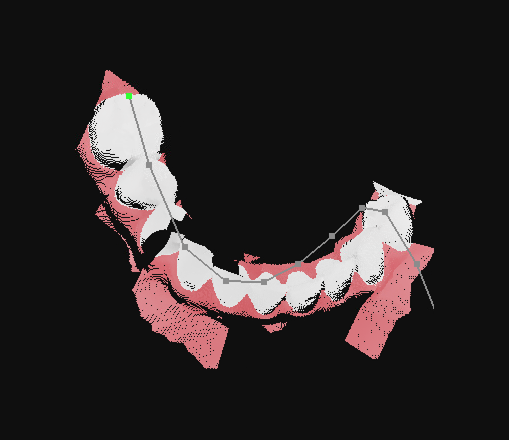}
&
\includegraphics[width=0.18\linewidth]{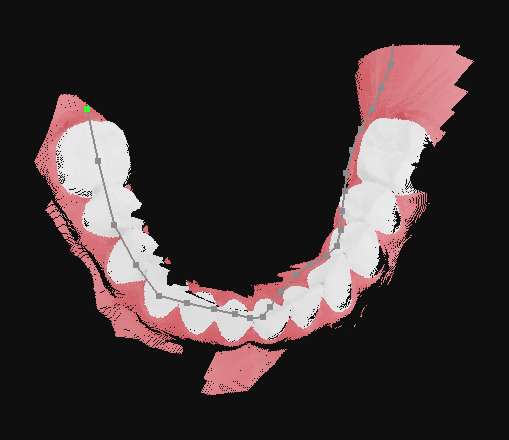}
&
\includegraphics[width=0.18\linewidth]{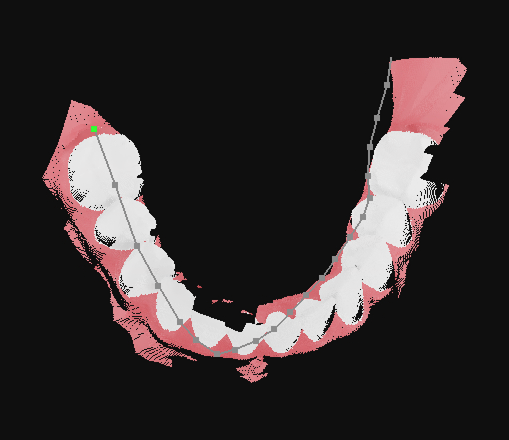}
&
\includegraphics[width=0.18\linewidth]{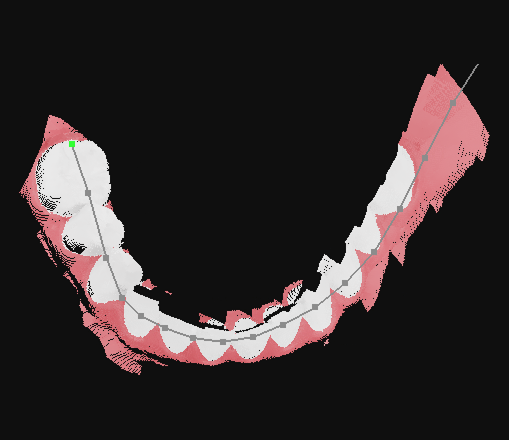}
&
\includegraphics[width=0.18\linewidth]{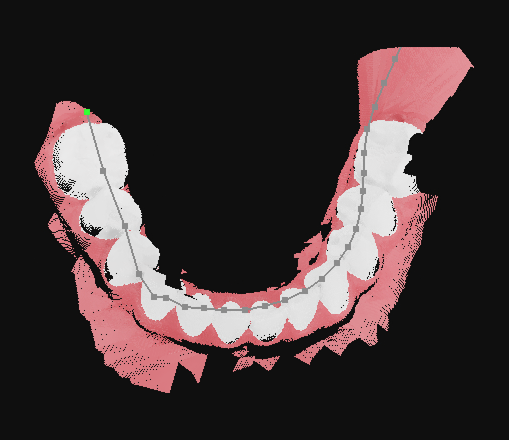} \\
\bottomrule
\end{tabular}
\end{table}

The qualitative examples in Table~\ref{tab:reward_ablation_qualitative} are
consistent with the aggregate metrics and termination diagnostics. Coverage
Completion Only produces incomplete and imbalanced accumulation, whereas the full
reward yields the most complete arch-level reconstruction. The leave-one-out
variants still recover substantial surface regions, but each loses a different
desirable behavior: balanced tooth-wise completion, stable scanner placement, or
compact execution. Overall, the results support that the full reward provides the
best balance between tooth-wise completion, stable scanner placement, and
efficient scan execution.
\section{Zero-Shot Hardware Deployment}
\label{app:hardware}

\subsection{Robot-Scanner Setup and Mount}
\label{app:hardware_setup}

\begin{figure}[htbp]
    \centering
    \includegraphics[width=\linewidth]{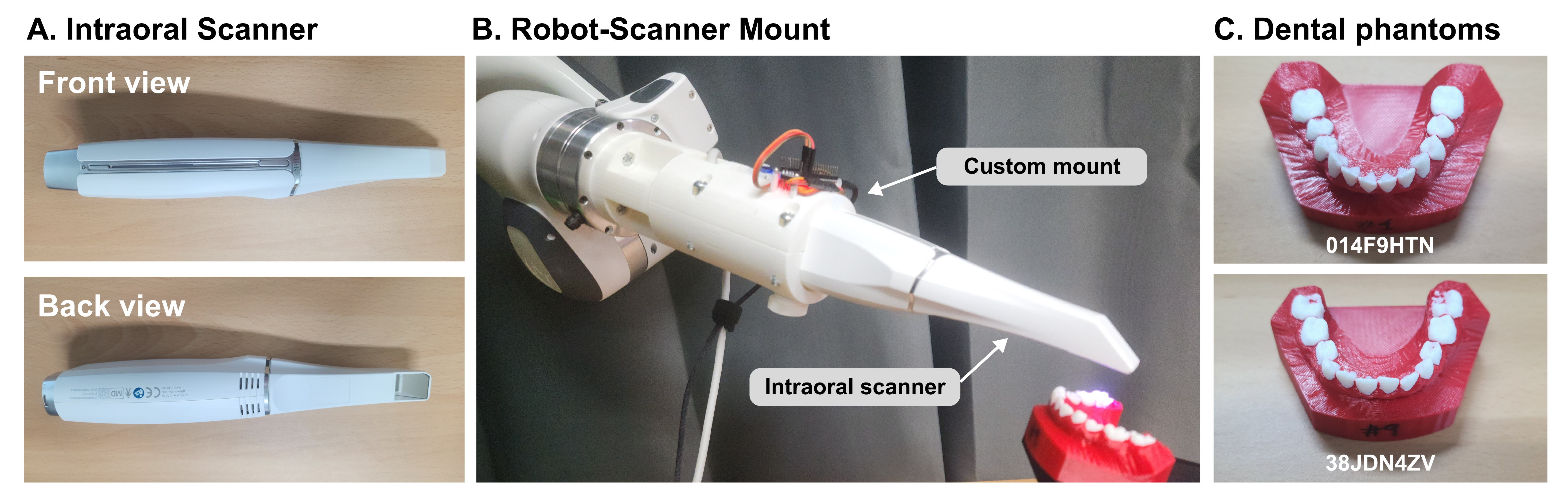}
    \caption{
        Hardware deployment setup. 
        (A) Front and back views of the intraoral scanner. 
        (B) The robot-scanner mount, showing the scanner rigidly attached to the robot end-effector via a custom 3D-printed part. 
        (C) Dental phantoms used for zero-shot hardware rollouts.
    }
    \label{fig:hardware_setup}
\end{figure}

As illustrated in Figure~\ref{fig:hardware_setup}, we deploy the learned policy on a physical robot-scanner setup consisting of a Franka Research 3 (FR3) arm and a Huvitz Lilivis SCAN intraoral scanner.
The scanner is rigidly attached to the robot end-effector using a custom 3D-printed mount.
The dental phantom is placed within a conservative Cartesian scanning region $\mathcal{W}_{\mathrm{hw}}$, defined in the robot base frame.
This region specifies the intended operating region for the phantom setup and should not be interpreted as the full reachable workspace of the robot.

To ensure consistent point-cloud accumulation during hardware rollouts, we estimate the static rigid transform $T^{T \leftarrow C}$ between the robot tool frame $T$ and the scanner camera frame $C$ via standard hand-eye calibration prior to deployment.
Using this fixed calibration, the local point cloud $P_t^C$ acquired by the scanner at timestep $t$ is transformed into the world frame $P_t^W$ as
\begin{equation}
P_t^W =
T^{W \leftarrow B}
T_t^{B \leftarrow T}
T^{T \leftarrow C}
P_t^C,
\end{equation}
where $T^{W \leftarrow B}$ is the static base transform, $T_t^{B \leftarrow T}$ is the robot end-effector pose obtained from forward kinematics at each control step, and $T^{T \leftarrow C}$ is the fixed hand-eye transform.

\subsection{Qualitative Zero-Shot Hardware Rollouts}
\label{app:real_rollouts}

We evaluate the trained policy on the physical robot-scanner setup without real-world fine-tuning, reward adaptation, or additional policy optimization.
At each control step, the policy receives the accumulated geometric memory and robot proprioception, and predicts a relative 6-DoF scanner motion for closed-loop scan control.
Representative simulation and hardware outcomes are shown in Table~\ref{tab:sim_real_qualitative}.

\begin{table}[!htbp]
\centering
\caption{
Representative simulation and hardware deployment outcomes on the target jaw.
}
\label{tab:sim_real_qualitative}
\small
\setlength{\tabcolsep}{3pt}
\renewcommand{\arraystretch}{0.95}
\begin{tabular}{@{}cc@{}}
\toprule
Simulation & Real-world deployment \\
\midrule
\includegraphics[height=0.23\textheight,keepaspectratio]{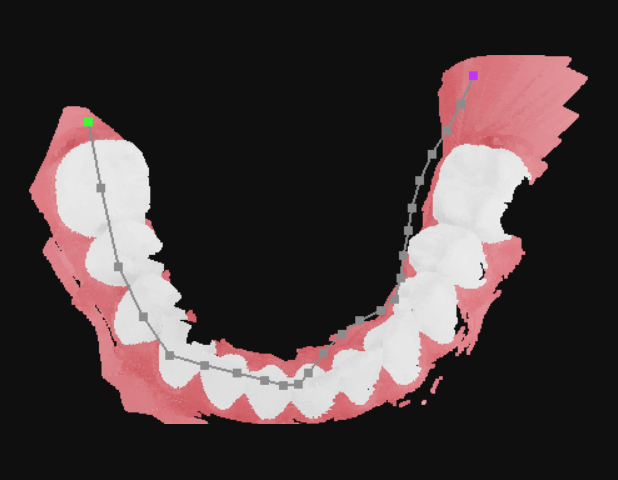}
&
\includegraphics[height=0.23\textheight,keepaspectratio]{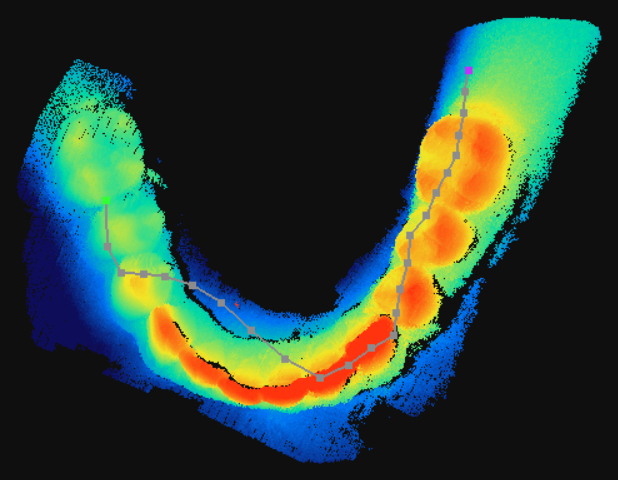} \\
\bottomrule
\end{tabular}
\end{table}

The hardware rollout maintains stable scanner motion around the target jaw and progressively accumulates visible tooth and adjacent-gingiva surfaces from multiple viewpoints.

\subsection{Deployment Safety and Limitations}
\label{app:hardware_safety}

The target scanner pose induced by the policy output is converted into robot joint commands using differential inverse kinematics.
During hardware execution, the robot-side controller enforces joint-position and joint-velocity limits.
Therefore, when a commanded motion would require a joint to move beyond its feasible range, the realized motion is limited by the corresponding joint boundary.
This limit handling is applied at the robot-control layer for hardware safety and is not provided as an additional policy input.

The hardware experiment is intended as a qualitative zero-shot feasibility demonstration under a phantom setup.
It is not a clinical in-mouth validation.
Remaining sim-to-real error sources include scanner noise, hand-eye calibration error, robot execution latency, and pose drift during point-cloud accumulation.
These limitations motivate future work on online pose correction, improved hand-eye calibration, and closed-loop registration using real scanner feedback.
